\documentclass[runningheads]{llncs}
\usepackage{graphicx}
\usepackage{mathtools}
\usepackage{amssymb}
\usepackage{bm}
\usepackage{bbm}

\usepackage{booktabs} 
\usepackage{color}
\usepackage{algorithm}  
\usepackage[noend]{algpseudocode}

\begin{document}
\title{Learning Differential Diagnosis of Skin Conditions with Co-occurrence Supervision using Graph Convolutional Networks}
\titlerunning{GCN-CNN}
%

\author{Junyan Wu\inst{1} \and 
Hao Jiang\inst{2} \and 
Xiaowei Ding\inst{1, \star} \and 
Anudeep Konda\inst{1} \and 
Jin Han\inst{1} \and 
Yang Zhang\inst{1} \and 
Qian Li\inst{1}} 

\institute{Voxelcloud Inc., Los Angeles, California, USA\\
\email{xding@voxelcloud.io}\and
Northeastern University, Shenyang, Liaoning Province, China}

\footnotetext[1]{$^{\!\!,\star}$ To whom correspondence should be addressed}

\authorrunning{ }
\titlerunning{ }

\maketitle              
\begin{abstract}
Skin conditions are reported the 4$^\text{th}$ leading cause of nonfatal disease burden worldwide. However, given the colossal spectrum of skin disorders defined clinically and shortage in dermatology expertise, diagnosing skin conditions in a timely and accurate manner remains a challenging task. Using computer vision technologies, a deep learning system has proven effective assisting clinicians in image diagnostics of radiology, ophthalmology and more. In this paper, we propose a deep learning system (DLS) that may predict differential diagnosis of skin conditions using clinical images. Our DLS formulates the differential diagnostics as a multi-label classification task over 80 conditions when only incomplete image labels are available. We tackle the label incompleteness problem by combining a classification network with a Graph Convolutional Network (GCN) that characterizes label co-occurrence and effectively regularizes it towards a sparse representation. Our approach is demonstrated on 136,462 clinical images and concludes that the classification accuracy greatly benefit from the Co-occurrence supervision. Our DLS achieves 93.6\% top-5 accuracy on 12,378 test images and consistently outperform the baseline classification network.

\end{abstract}
\section{Introduction}
Skin problems and conditions are common health concerns and their diagnostics are largely based on visual clues. 
According to \cite{lim2017burden}, 27\% of the U.S. population were seen by a physician for skin disease in 2013 and the affected individuals averaged 1.5 skin diseases. Diagnosing and treating skin conditions remains a challenge as a diverse set of skin diseases, with over 3000 entities identified in the literature \cite{segre2006epidermal}, need to be differentiated by dermatologists who are in significant shortage relative to the rising demand. As a first step, deriving a group of possible causes from visual impression is deeply rooted in clinical practice, therefore often refereed to as ``differential diagnosis'' (see \cite{ashton2014differential}). 

Recent advances in computer vision promise an accessible and reliable solution to differential diagnosis on clinical skin images. Previous work has demonstrated the efficacy of Covolutional Neural Networks (CNN) powering decision support systems to radiologists, ophthalmologists and pathologists. In dermatology, dermoscopy images has attracted much attention in which the image modality is more standardized and target labels are multiple magnitudes less than the number of skin conditions. Recently, more efforts are cast on clinical images in a direct effort to target multiple skin conditions, for example in \cite{liu2019deep}, a deep learning system(DLS) was trained to distinguish 26 disease classes. In their work, authors misrepresented differential diagnosis as a multi-class classification problem therefore inherently undermined the interpretability of predictions and correlation between labels. \cite{esteva2017dermatologist} took on a dataset of 129,450 clinical images consisting of 2032 skin conditions, however, it overlooked the differential diagnosis problem and evaluated their DLS mostly on binary classification tasks, \emph{i.e.} cancer versus non-cancer. 

In this paper, we propose a deep learning system (DLS) that may predict differential diagnosis of skin conditions using clinical images. Our DLS formulates the differential diagnostics as a multi-label classification task over 80 conditions when only incomplete image labels are available. We tackle the label incompleteness problem by combining a classification network with a Graph Convolutional Network (GCN) that characterizes label co-occurrence and effectively regularizes it towards a sparse representation. Our approach is demonstrated on 136,462 clinical images and concludes that the classification accuracy greatly benefit from the co-occurrence supervision. 

Our GCN-CNN approach highlights three major advantages:
\begin{itemize}
    \item By introducing co-occurrence supervision by means of GCN layers, we effectively handle correlated image labels even when annotations are incomplete.
	\item Our approach is end-to-end trainable and readily applicable to any CNN architecture.
	\item GCN can be flexibly initialized by either empirical or expert-provided inputs that may adapt well per applications.
\end{itemize}

Finally, we evaluate our DLS on 12,378 user taken images acquired through a telehealth platform. To the best of our knowledge, this is the first study to investigate the performance of a DLS to differentiate skin conditions outside clinic. We report a top-5 accuracy of 93.6\% on test images and argue for the value of DLS in extending the reach of dermatological expertise with tremendous accessibility and accuracy.

\section{Related Work}
Graph Convolutional Networks (GCNs) were first introduced in \cite{kipf2016semi}. In its original application, \emph{i.e.} the problem of nodes classification, only a small subset of nodes had their labels available. By introducing a fast approximation to spectral graph convolutions, labels to unknown nodes can be effectively learned in a semi-supervised manner as information from labeled nodes propagates through GCN layers.

In \cite{chen2019multi}, ML-GCN was proposed for multi-label classification task. Different from \cite{kipf2016semi}, a graph structure was constructed from data and ML-GCN may directly incorporate a representation learned from a convolutional network. The graph structure is a directed graph over object labels and an edge of ``$\text{Label}_\text{i} \rightarrow \text{Label}_\text{j}$'' means when $\text{Label}_\text{i}$ is present, $\text{Label}_\text{j}$ is likely to be present too.

Formally, the output of ML-GCN, \emph{i.e.}, $\bm{W} = \{\bm{w}_i\}_{i=1}^C$, parameterizes a mapping from feature vectors, learned from a conventional convolution network, to $C$ labels. As for the final prediction, a CNN-based model learns an image representation $\bm{x}$ and the predicted scores $\bm{y}$ can be derived as,
\begin{align}
    \text{CNN}\quad    & \widehat{\bm{y}} = f(\bm{W}\cdot\bm{x})\\
    \text{ML-CNN}\quad         & \widehat{\bm{y}} = f(\bm{\tilde{W}(D, Z)}\cdot\bm{x}),
\end{align}
where $D$ is the directed graph derived from correlation, $Z=\{z_i\}_{i=1}^C$ is a set of semantic embeddings to each label, both of which are predetermined therefore fed into GCN based classifier $\bm{\tilde{W}}$. As authors claimed, $\bm{\tilde{W}}$ benefits from both conditional dependence characterized in $\bm{D}$ and semantic proximity embedded in $\bm{Z}$.

\section{Method}

\begin{figure}[!htb]
\includegraphics[width=\textwidth]{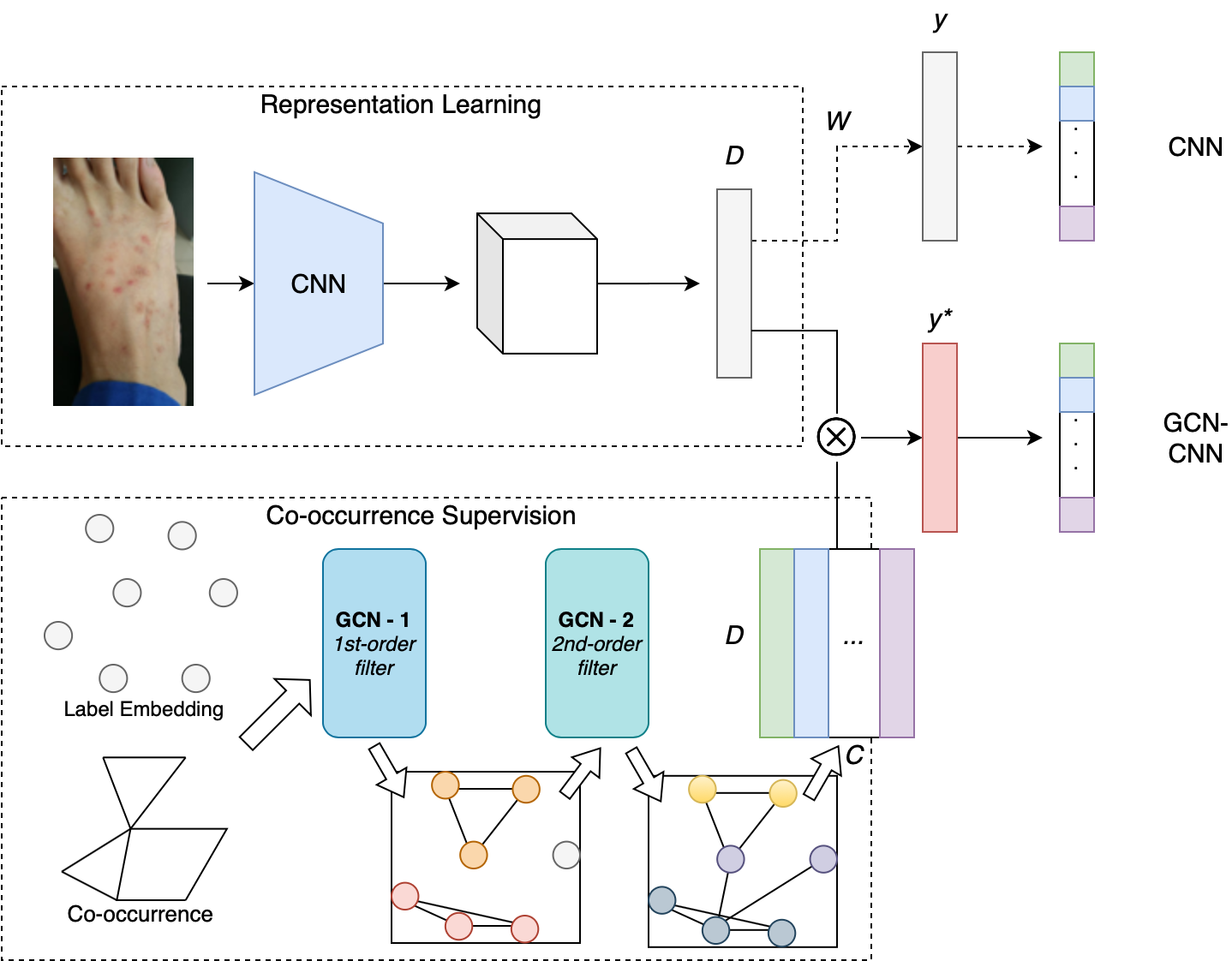}
\caption{Overview of GCN-CNN: the GCN branch propagates label co-occurrence and semantic embedding. A trainable representation network has its feature vectors dot product with GCN output and generate final predictions.} 
\label{fig:GCNCNN}
\end{figure}

Our GCN-CNN approach naturally extends ML-GCN and tailor it specifically to the differential diagnostics problem. The overall framework of GCN-CNN is presented in Fig~\ref{fig:GCNCNN}. An undirected graph replaces the directed graph in \cite{chen2019multi} given the symmetry of conditional probability when two conditions are discerned. We leverage spectral graph convolution in \cite{kipf2016semi} between GCN layers. Finally, we demonstrate and evaluate the graph construction empirically from label co-occurrence and by medical expert.

Our GCN branch consists of two graph convolutional (GC) layers. We follow the work of \cite{kipf2016semi} and use the $k-$th order filter of spectral graph convolution that propagates the neighboring nodes up to $k$ steps. The first GC layer (GCN-1) takes order 1 and convolves directly on neighboring nodes that is equivalent to a $1-$st order spectral filter, similarly, the second layer (GCN-2) takes the same order and extends to indirect neighbors as a $2-$nd order filter on the original graph. Except for computational advantage which reduces the graph convolution complexity to linear to the number of edges, this characterization avoids over smoothing label nodes by a deep GCN.

During the training time, we empirically estimate a co-occurrence graph using only training data. An undirected graph $\mathcal{G} = \left\{\mathcal{V},\mathcal{E}\right\}$ encodes the conditional dependency between image labels, \emph{i.e.} skin condition, and implicitly supervises the classification task. Node representations in $\mathcal{V}$ embed semantic meaning to labels which complements $\mathcal{G}$ particularly when one label includes another as a substring. As for edges, $e_{i,j} \in \mathcal{E}$ is sparsely constructed such that 

\begin{equation}
    e_{i,j} = \mathbbm{1}(\frac{C(i, j)}{C(i) + C(j)} \ge t) ,
    \label{eqn:graphedge}
\end{equation}
where $C(i, j)$ is the number of images that have both label $i$ and $j$, $C(i)$ and $C(i)$ are the total number of images in class $i$ and $j$ respectively. In our experiment, a differential graph is also constructed with domain knowledge by board certified dermatologists. We ask two dermatologists to provide overlapped differential diagnoses groups, as many as possible, and connect an edge when two labels appear in at least one differential group by both dermatologists.

Since our approach is end-to-end trainable, we simply use multi-label cross entropy as loss function to the classification task. 

\section{Experiments}
\subsection{Dataset}
A total of 136,462 user taken images were used for training and testing. In our dataset, the images are directly acquired by end users on a telehealth platform operating in China. All images are extracted randomly from consultations to physicians that involves skin problems. We split this dataset into a training set of 124,084 images and a test set of 12,378 images and conduct annotation differently.

For the training set, we obtain single reader annotation by randomly selecting a dermatologist from a group of 25. Due to the cost of annotating medical images and limited expertise available to the task, single reading on the image is commonly used in practice. To each dermatologist, we present the image and ask for their impression of skin conditions relative to the symptoms that manifests, as many as possible. Upon finish, 81.7\% of training images carry a single label which outnumber doubly labeled images (15.5\%) and triply labeled images (2.8\%).

Multi-reader annotations were collected and aggregated on the test set. At least two dermatologists were blindly and independently presented with the same images and asked to conduct annotation the same as training. As the first two dermatologists fail to converge, a third dermatologist is involved and final labels are determined by majority voting. In contrast to the training set, testing images have 46.0\% singly labeled, 38.1\% doubly labeled and 12.7\% triply labeled. The label distribution is significantly different between training and testing and it highlights the label incompleteness issues commonly encountered on single reader annotations.

Finally, we select the top 80 frequent conditions and a complete list is available in Appendix A. Differential graph contributed by domain experts takes into account lesion morphology, configuration and distribution, more details are provided in Appendix B.

\subsection{Experiment Setting}

The node representation of our GCN takes dimensions of 700 (GCN-0), 1024 (GCN-1) and 2048 (GCN-2). The initial label embedding input at GCN-0 utilizes BioSentVec (\cite{chen2019biosentvec}) specifically trained on biomedical corpus. We investigate two undirected graph construction: empirical label co-occurrence and differential graph by domain experts.

All input images are downsized to $448 \times $448 and we use a Resnet-101\cite{he2016deep} as the classification backbone. A linear layer was added after FC-2048 to conduct dot product between image features and GCN node features.

During training time, CNN backbone is first trained with 300 epochs at an initial learning rate of $0.1$ with step decay. Thereafter, we randomly initialize GCN branch and train the whole GCN-CNN architecture end-to-end for another 300 epochs at a learning rate of 0.0003. 

To formally evaluate our approach, we consider Resnet-101 alone as the baseline and benchmark our method against Li, Y \textit{et al.}\cite{li2017improving}, which improved multi-label classification by means of a novel loss function for pairwise ranking. More efforts focus on the GCN branch as multiple constructions are formally compared.

\subsection{Evaluation Metrics}

Top 1/3/5 accuracy and mAP across all labels are major metrics of interest in evaluation. Additionally, we report Hamming Loss, Ranking Loss and Ranking One Error that are particularly suited for multi-label classification problems. Formal definitions are available in Appendix D. 

It is worth mentioning that top-$n$ accuracy considers a prediction true positive when the top-$n$ predictions overlap with GT label set. Therefore, it characterizes the relevance of model prediction rather than the completeness of differential label sets. Authors recommend multi-label metrics since they complement top-$n$ metrics and hint on the comprehensiveness of model predictions while adjusting for the prediction ranking.

\section{Results}
\subsection{Performance Gain on Co-occurrence Graph}
As shown in Table~\ref{tab:tabel1}, GCN-CNN with empirical co-occurrence graph outperforms its competitors by all measures. It also predicts differential diagnoses more accurately and comprehensively based on multi-label classification metrics in Table~\ref{tab:tabel2}. Random graph initialization leads to inferior performance to baseline, not surprisingly, given the limited depth of GCN constrains its representativeness of label dependency.

\begin{table}\centering
\caption{Performance Comparison: Classification Metrics}\label{tab:tabel1}
\begin{tabular}{cccccccc}
\toprule
 ~Method~ &  ~top1 acc~ &  ~top3 acc~ &  ~top5 acc~ &  ~mAP~ 
\\ \midrule
Resnet-101 & 0.682 & 0.866 &  0.918 & 0.5067\\
Li, Y \textit{et al.}~\cite{li2017improving} & 0.689 & 0.866 & 0.916 & 0.498\\
Ours (random graph)  & 0.653 & 0.855 & 0.912 & 0.469\\
Ours (co-occurrence graph) & \bf 0.703 & \bf 0.885 & \bf 0.936 & \bf 0.546 \\
\bottomrule
\end{tabular}
\end{table}

\begin{table}\centering
\caption{Performance Comparison: Multi-Label Metrics}\label{tab:tabel2}
\begin{tabular}{cccccccc}
\toprule
Method &  ~Hamming Loss~ &  ~Ranking Loss~ &  ~Ranking One Error~
\\ \midrule
Resnet-101 &  0.164 & 0.475 & 0.326\\
Li, Y \textit{et al.}~\cite{li2017improving} & 0.171 & 0.492 & 0.341\\
Ours (random graph)  & 0.274 & 0.496 & 0.340\\
Ours (co-occurrence graph) & \bf 0.093 & \bf 0.456 & \bf 0.287\\
\bottomrule
\end{tabular}
\end{table}

\begin{figure}[!ht]
\includegraphics[width=\textwidth]{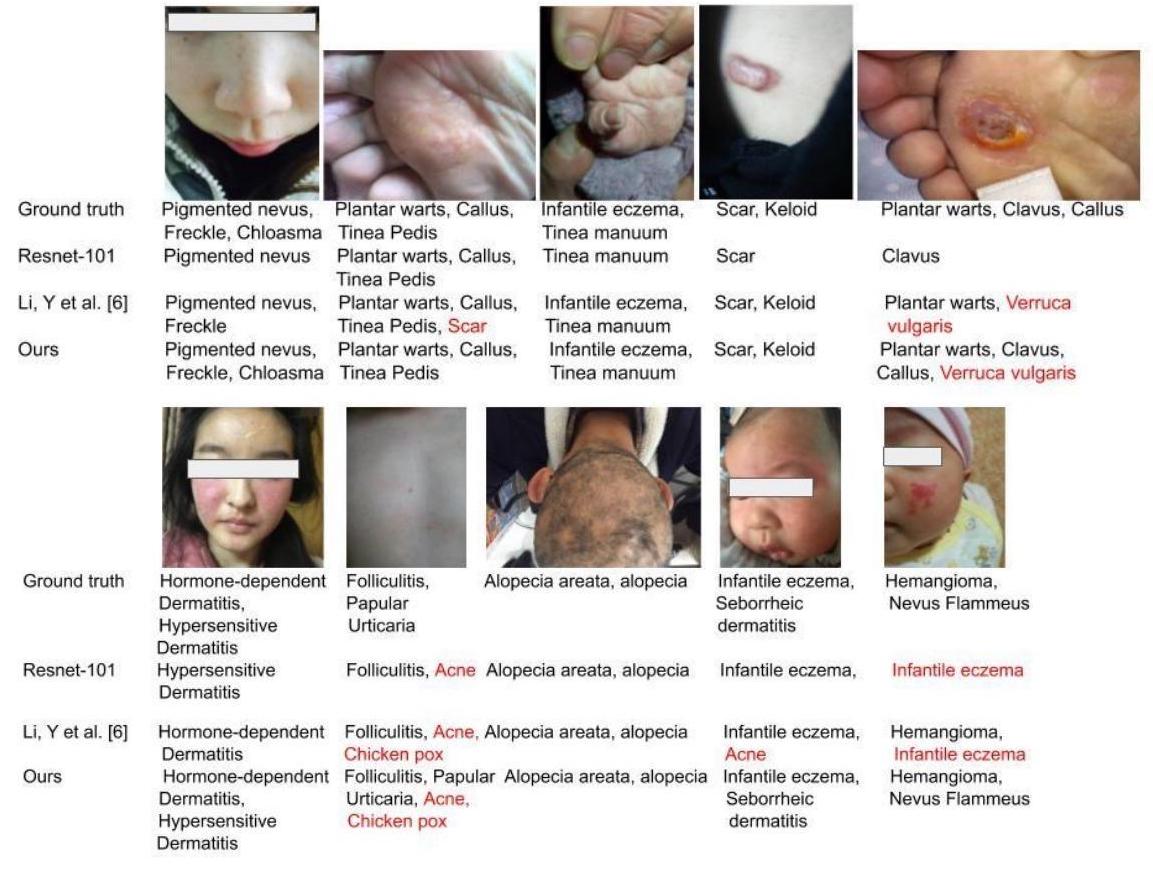}
\caption{Sample predictions: false positives are colored red.} \label{fig2}
\end{figure}

We exemplify model predictions in Figure~\ref{fig2}. The baseline Resnet makes fewer predictions given the imbalanced labels between training and testing. The label incompleteness also deteriorates the performance of Li, Y \textit{et al.} which is incapable of leverage co-occurrence information effectively. The number of false positive predictions by Li, Y \textit{et al.} also increases as label ranking is adjusted based upon incomplete supervision.

Given the sparsity of co-occurring labels in our training data, we did not investigate the sensitivity of model performance to the choice of threshold $t$ in Eq.~\eqref{eqn:graphedge}.

\subsection{Graph Output from GCN}

\begin{figure}[!htb]
\includegraphics[width=\textwidth]{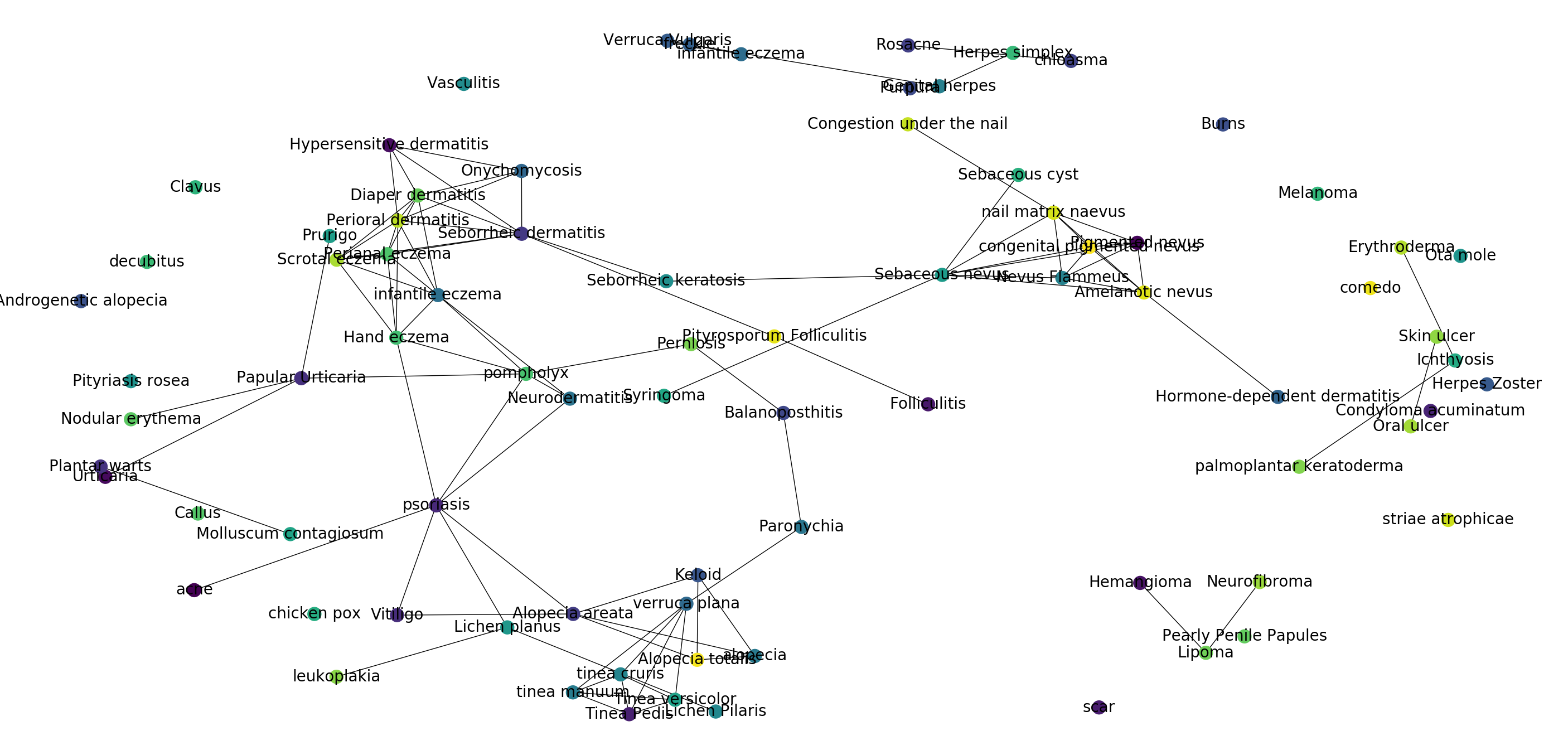}
\caption{Nodes proximity at GCN-0: BioSentVec Embeddings.} \label{fig3}
\end{figure}

\begin{figure}[!htb]
\includegraphics[width=\textwidth]{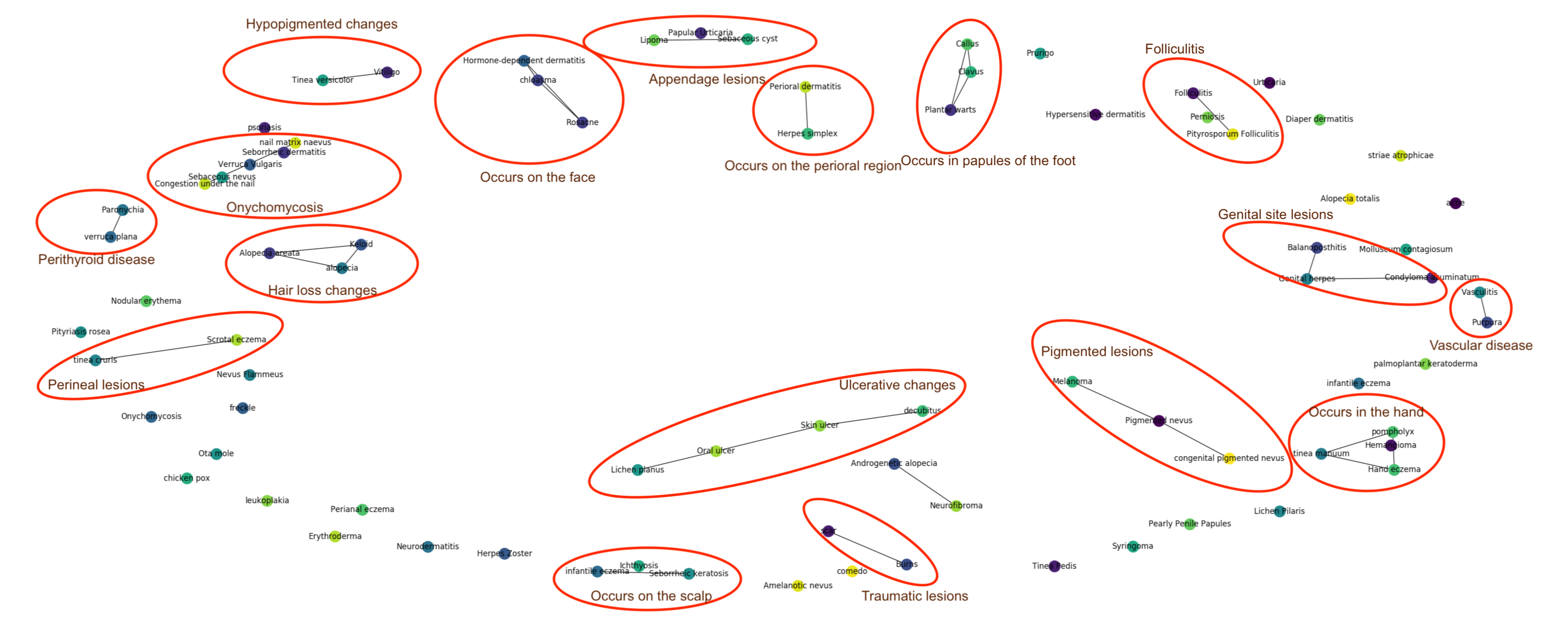}
\caption{
Nodes proximity at GCN-2: isolated clusters are highlighted and interpreted by dermatologists.} \label{fig4}
\end{figure}

As a byproduct of our GCN banch, the proximity of nodes after GCN-2 may hint on a refined label dependency. We consider the difference between GCN-0 nodes proximity and GCN-2 nodes proximity as the learned label dependency. In Figure~\ref{fig3}, BioSentVec encoded label proximity is mainly driven by their semantic meaning. However in Figure~\ref{fig4}, label proximity has reduced to isolated clusters that highly correlate with differential groups, e.g. Appendage lesions, Perithyroid disease, Ulcerative changes, etc. This observation explains how GCN-CNN has improved classification accuracy and introduces extra interpretability of the system.

\subsection{Differential Graph by Domain Experts}
We also evaluate an undirected graph construction by domain experts.
As shown in Table~\ref{tab:performance2}, differential graph achieves comparable performance against empirical co-occurrence graph. Marginal gains are observed on top 1, 3 and 5 accuracy but none is statistically significant. Considering the laborious task to construct such graph when the label set grows, empirical co-occurrence graph is an strongly recommended.

\begin{table}\centering
\caption{Performance comparison for different graph construction}\label{tab:performance2}
\begin{tabular}{ccccc}
\toprule
\bf ~Method~ & \bf ~top1 acc~ & \bf ~top3 acc~ & \bf ~top5 acc~ & \bf ~mAP~  
\\ \midrule
Ours (co-occurrence graph) & 0.703 & 0.885 & 0.936 & \bf 0.546  \\
Ours (knowledge graph)  & \bf 0.708 & \bf 0.891 & \bf 0.940 & 0.545 \\
\bottomrule
\end{tabular}
\end{table}

\section{Conclusion}
We introduce label co-occurrence supervision via a GCN branch for the problem of differential diagnosis of skin conditions. Our approach has significantly improved classification accuracy and completeness even when trained on incomplete labels that are commonly seen in medical imaging applications.
By testing on user taken images of skin issues, we report a top-5 accuracy of 93.6\%. This deep learning system is promising to be used as clinical decision support to medical professional with limited training in dermatology, as well as an accessible self-diagnostic tool directly to consumers.

Besides, the GCN branch leads to explainable visualization of label proximity that can be readily utilized for interpretation and debugging. Moreover, our GCN-CNN approach is end-to-end trainable, may adapt to any classification backbone and add zero effort during inference time. Furthermore, the GCN branch may extend to a diversity of features, e.g. patient demographics and medical history, therefore leverage multi-modal information in a classification task.

\bibliographystyle{splncs04}
\bibliography{ref}

\end{document}